\documentclass[letterpaper, 10 pt, conference]{ieeeconf}  

\IEEEoverridecommandlockouts                              

\usepackage{amsfonts}
\usepackage{amsmath}
\usepackage{amssymb}
\usepackage[]{todonotes}
\usepackage{graphicx}
\usepackage{hyperref}
\usepackage{caption}
\usepackage{subcaption}

\usepackage{tabularx}
\usepackage{booktabs}
\usepackage{balance}
\usepackage{color,soul}
\usepackage{cite}
\usepackage{bm}
\usepackage{siunitx}
\title{\LARGE \bf
Flexible Gear Assembly With Visual Servoing and Force Feedback
}

\author{Junjie Ming, Daniel Bargmann, Hongpeng Cao, and Marco Caccamo 
\thanks{Junjie Ming and Daniel Bargmann are with the Fraunhofer Institute for Manufacturing Engineering and Automation, Stuttgart, Germany. {\tt\small junjieming@gmail.com, danb@ipa.fraunhofer.de}}%
\thanks{Hongpeng Cao is with the School of Engineering and Design, Technical University of Munich (TUM), Munich, Germany. {\tt\small cao.hongpeng@tum.de}}
\thanks{Marco Caccamo is with the School of Engineering and Design, Technical University of Munich (TUM), Munich, Germany and Munich Institute of Robotics and Machine Intelligence, Technical University of Munich (TUM), Munich, Germany. {\tt\small mcaccamo@tum.de}}%
\thanks{A video showing the gear assembly process with the proposed approach is available at \url{https://www.youtube.com/watch?v=5BDMzEspp_M&ab_channel=HongpengCao}.}
}

\begin{document}

\maketitle
\thispagestyle{empty}
\pagestyle{empty}

\begin{abstract}
Gear assembly is an essential but challenging task in industrial automation. This paper presents a novel two-stage approach for achieving high-precision and flexible gear assembly. The proposed approach integrates YOLO to coarsely localize the workpiece in a searching phase and deep reinforcement learning (DRL) to complete the insertion. Specifically, DRL addresses the challenge of partial visibility when the on-wrist camera is too close to the workpiece. Additionally, force feedback is used to smoothly transit the process from the first phase to the second phase. To reduce the data collection effort for training deep neural networks, we use synthetic RGB images for training YOLO and construct an offline interaction environment leveraging sampled real-world data for training DRL agents. We evaluate the proposed approach in a gear assembly experiment with a precision tolerance of 0.3~mm. The results show that our method can robustly and efficiently complete searching and insertion from arbitrary positions within an average of 15 seconds.
\end{abstract}
\section{INTRODUCTION}
\label{sec:introdution}

\begin{figure}[ht]
    \vspace*{0.17cm}%

    \centering
    \includegraphics[width=0.8\columnwidth]{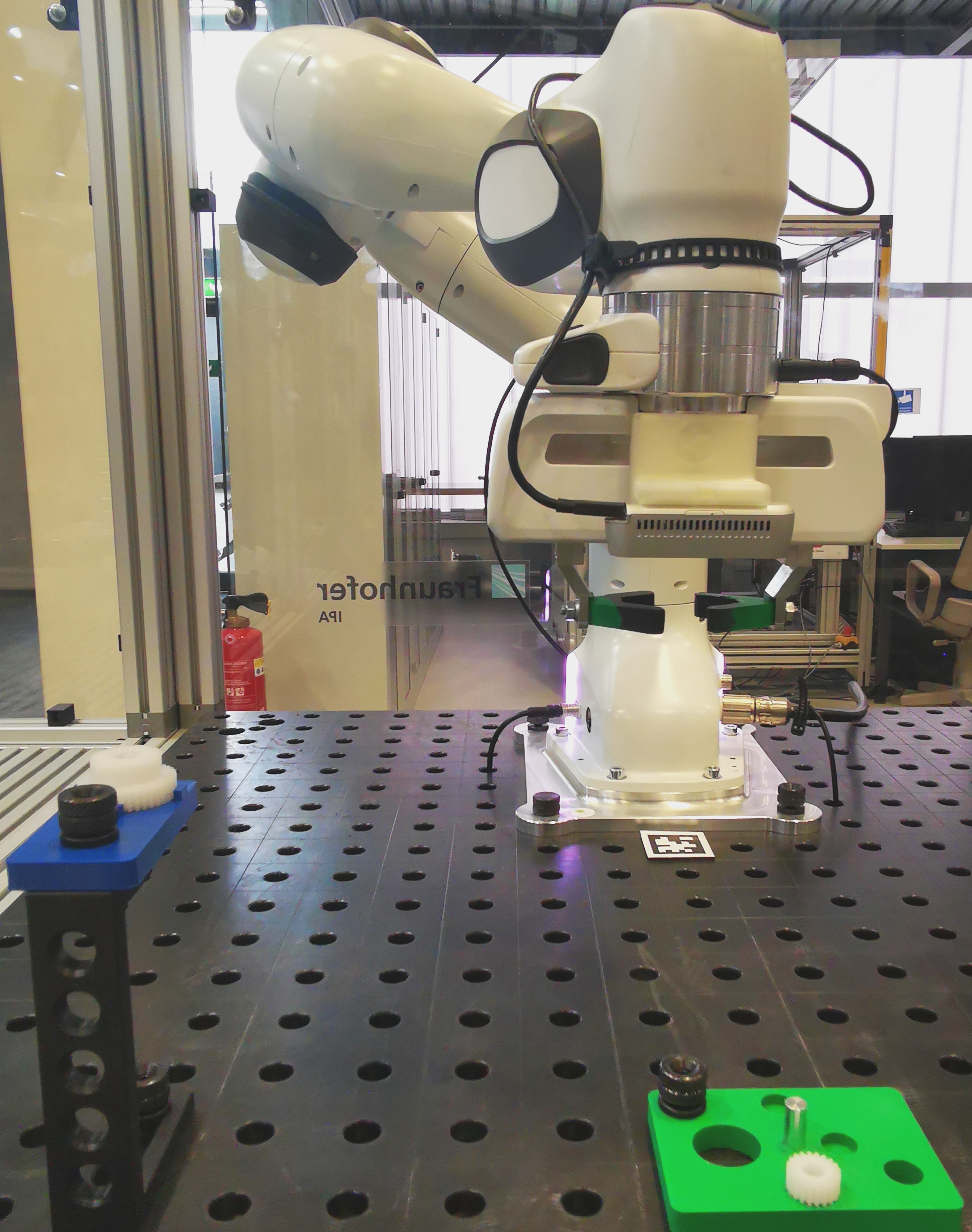}
    \caption{The figure provides an overview of the gear assembly task, with the objective of grasping a gear from the placement tray, observing the environment using the on-wrist RGBD camera, and inserting the gear held by the robot onto the peg using visual servoing to complete the assembly.}
    \label{fig: scenario}
    \vspace*{-0.27cm}%

\end{figure}

\begin{figure}[h]
    \vspace*{0.17cm}%

    \centering
    \subfloat[Assembly misalignment errors after the first stage]{\includegraphics[width=0.46\columnwidth]{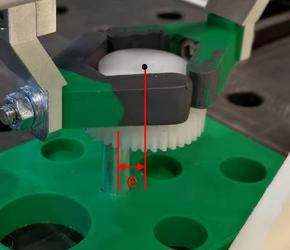}\label{fig:offset}} 
    \hspace{0.05cm}
    \subfloat[View of the robot camera during assembly phase]{\includegraphics[width=0.46\columnwidth]{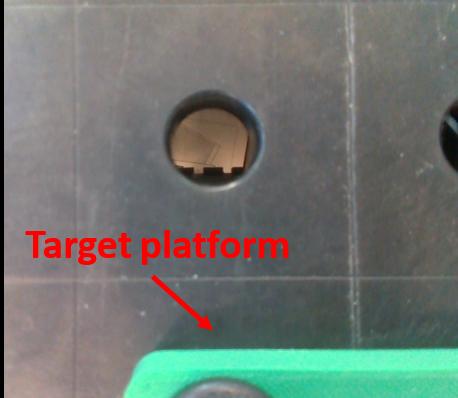}\label{fig:partial_obs}}
\caption{The figure shows the assembly status after localization from the first stage.}
\vspace*{-0.27cm}%
\end{figure}

High-precision and flexible gear assembly have been essential in industrial automation \cite{li2019survey}. Typically, a gear assembly process involves searching and insertion \cite{sharma2013intelligent}. To complete these two procedures, conventional gear assembly relies on tedious position teaching, which has the limitations of time cost and low flexibility \cite{biggs2003survey}. Recent work leverages visual information \cite{litvak2019learning, huang2017vision, 9646215, de2018integration} and force/torque (FT) feedback \cite{hou2020data, inoue2017deep, luo2018deep, hou2020fuzzy, luo2019reinforcement, oikawa2021reinforcement} in robotic assembly to avoid the reliance on tedious position teaching, achieving promising performance. 

In this work, we aim to develop an efficient solution to realize the part searching and insertion for high precision and flexible gear assembly task, as shown in Figure \ref{fig: scenario}. The assembly scenario consists of a placement tray with an initial gear, a Frank Emika Panda Robot, and the target platform mounted with a gear near the peg to be assembled. The robot is equipped with an RGBD camera on the wrist to observe the environment and a force sensor on the gripper to detect contact. This assembly process aims to grasp a gear from the placement tray and insert it onto the peg to finish the assembly with the gear on the platform under the precision tolerance of \SI{0.3}{\mm}. We simplify the grasping by initializing the gear with a known position on the placement tray and pre-program a fixed motion for grasping, as the assembly is the main focus of this work.

We propose a vision-based two-stage approach with force feedback to solve the above gear assembly problem. Specifically, in the first stage, we use YOLO \cite{yolov3} with depth information to roughly localize the object. With the coarse position, we can efficiently drive the robot using position control to reach the neighbor of the peg. We use force feedback to determine whether the gear touches the peg or not for deciding the transition between the two stages. 

As shown in \ref{fig:offset}, after the first stage, the centers of the gear and peg are not aligned perfectly due to the perception error. Meanwhile, the camera loses the direct view of the peg, which imposes the challenges of predicting or measuring the relative position of the peg. We propose a novel solution to this challenge by training deep reinforcement (DRL) agents to learn the underlying relative positions from partially visible target platforms in RGB images, as shown in \ref{fig:partial_obs}. Through interaction with the environment, DRL learns to output movement setpoints for position control to complete the insertion. We explore DQN \cite{mnih2015human} algorithm with discretized action space and PPO \cite{ppo} with continuous action space for the insertion to study the potential of the proposed method. 

To reduce the data collection effort for training deep neural networks, we use synthetic RGB images for training YOLO and constructing an offline interaction environment leveraging sampled real-world data for training DRL agents. The real-world assembly experiment shows that the proposed approach can achieve high robustness and efficiency when tested in the gear assembly task from 100 different starting points.

To summarize our contributions, we propose a novel two-stage approach by integrating YOLO and deep reinforcement learning with force feedback for achieving high-precision and flexible gear assembly. We simplify the training of YOLO using synthetic images and ease the training of DRL agents with a real-images constructed offline training environment. For the insertion, we explore DQN and PPO to effectively address the challenge of partial visibility when the on-wrist camera is too close to the target. At last, we evaluate the proposed approaches in a real-world assembly experiment under varying positional conditions.

This work is structured as follows: Section \ref{sec:introdution} describes the background for the gear assembly task and an overview of the proposed solution. Section \ref{sec:related work} summarizes the related work. The proposed approach is introduced in detail in Section \ref{sec:methods}. The experimental setup and the evaluation of the proposed approach are discussed in Section \ref{sec:experiment}. Section \ref{sec:conclusion} concludes the work and gives a brief outlook on future work.
\section{Related work}
\label{sec:related work}
Robotic assembly is a challenging task in industrial automation. Related work leverages visual information \cite{litvak2019learning, huang2017vision, 9646215, de2018integration} and force/torque (FT) feedback \cite{hou2020data, inoue2017deep, luo2018deep, hou2020fuzzy, luo2019reinforcement, oikawa2021reinforcement} in robotic assembly to avoid the reliance on tedious position teaching, improving assembling reliability and flexibility. 

Vision-based approaches focus on pose detection for part searching, and recent work mainly aims to improve the detection accuracy \cite{litvak2019learning, huang2017vision, de2018integration}. In \cite{litvak2019learning}, a CNN-based algorithm is proposed to estimate the poses of the part from depth images and apply pose refinement to improve the estimation accuracy. \cite{9646215} detects the pose of the part from the 3D point cloud and gets more accurate pose estimation through refinement using the iterative close point (ICP) algorithm. The pose detection can also be simplified by detecting a checkerboard attached near the part \cite{litvak2019learning}. 

Once the poses of the parts are available, the assembly motion can be generated using a motion planer based on kinematics and geometry prior-knowledge \cite{chitta2012moveit}. The movement can be realized using position control. However, the quality of the vision-based approaches is highly dependent on the detection accuracy. Furthermore, it can be affected by the uncertainties presented in the insertion phase due to contact between different parts \cite{li2019survey}. This challenge motivates using force/torque (FT) sensors to provide contact information for the robot to deal with the uncertainties in the assembly process.

Directly deriving relative positions from F/T signals is challenging, as the contact dynamics model is complex \cite{oikawa2021reinforcement}. Deep reinforcement learning (DRL) holds the promise of completing the insertion by learning from interactions with the environment, where the dynamic model is not needed. Recent work \cite{hou2020data, inoue2017deep, luo2018deep, hou2020fuzzy, luo2019reinforcement} train DRL agents to map the F/T signals with other observations, for instance, the pose of the part, directly to the actions that drive the robot, achieving promising performance. In contrast to outputting action, \cite{oikawa2021reinforcement} proposes utilizing DRL to select stiffness matrix in admittance control to induce the motion that can complete insertion. 

Our work integrates YOLO to coarsely localize the target in the parts searching phase and deep reinforcement learning to align parts with finishing the insertion. Different from the related work that trains DRL to interpret relative positions from F/T signals for insertion, the novelty of our work lies in learning the underlying relative positions from partially observed target platforms in RGB images.

\section{Methods}
\label{sec:methods}
\begin{figure*}[t]
    \vspace*{0.17cm}%

    \centering
    \includegraphics[width=0.95\textwidth]{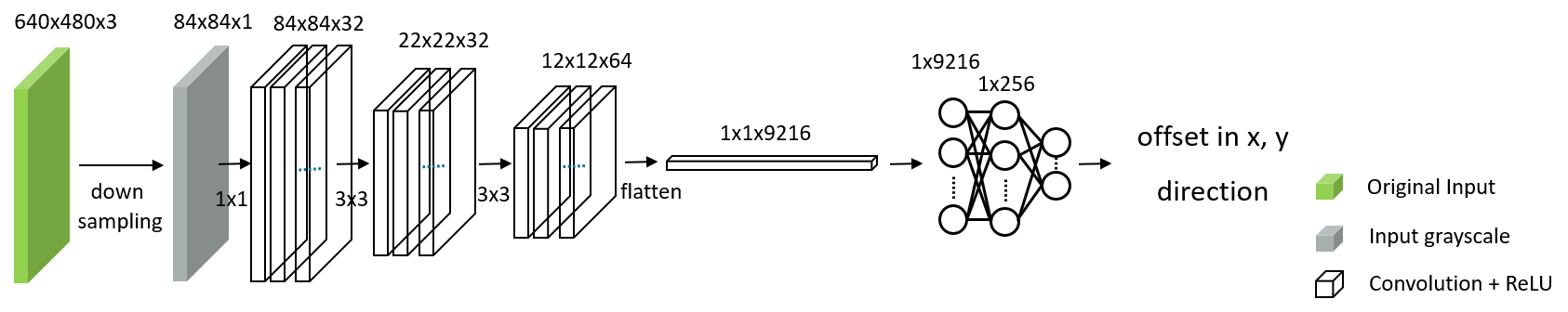}
    \caption{The plot illustrates the Convolutional Neural Network (CNN) utilized for extracting features from image input and outputting actions for the DQN and PPO agents.}
    \label{fig: network}

    \vspace*{-0.27cm}%

\end{figure*}
In this section, we introduce a vision-based two-stage approach with the assistance of force feedback to address the gear assembly problem. Specifically, in the first stage, we use YOLO \cite{yolov3} with depth information to roughly localize the object. In the second stage, we train deep reinforcement learning agents to mitigate the inaccuracies from the first stage and finish the gear assembly by utilizing force feedback. 

\subsection{Target localization}

The first stage of our assembly task involves localizing the target platform and bringing the end-effector to the center of the peg. We utilize an RGBD camera to obtain RGB and depth images of the assembly environment and apply the YOLO algorithm \cite{yolov3} on the RGB image to detect the 2D bounding box of the target platform. The depth image is then used to estimate the 3D coordinates $\{x_{pf},~y_{pf},~z_{pf}\}$ of the center of the platform with respect to the camera center, as in \cite{YOLO3_PLUS}. Specifically, we convert the pixel coordinates of the bounding box to 3D coordinates using the camera's intrinsic parameters and utilize the depth values along with the converted 3D coordinates to obtain the 3D position of the target platform. The 3D coordinates of the targeting peg $\{x_p,~y_p,~z_p\}$ are calculated by subtracting the relative distance from the detected center of the platform $\{x_{pf},~y_{pf},~z_{pf}\}$. Using position control, we then navigate the end-effector to $\{x_p,~y_p,~z_p\}$ and move the gripper downward along the z-axis. The first stage is deemed complete if the force detected in the z-axis exceeds the predetermined threshold $f_z$.

It should be noted that the localization method in the first stage may not align the center of the peg and gear perfectly due to the inaccuracies of the detected bounding box, resulting in an alignment error, as shown in Fig. \ref{fig:offset}. In the next section, we introduce training reinforcement learning agents to address the alignment error and complete the assembly task.

\subsection{Gear assembly}
To address the alignment error, we train DRL agents to generate $\{x,~y\}$ set-points and use the position control to navigate the end-effector. Meanwhile, we apply a minor $z$ position control to attempt downward insertion. Once the centers of the gear and peg are aligned, the gear can drop onto the peg, and the alignment process is completed. The gear is then further lowered using only $z$ control until it reaches the specified threshold $h_t$. Finally, a rotation along the $z$ axis is applied to complete the assembly process, resulting in the gear being successfully assembled with another gear mounted on the platform.

\subsubsection{Centers alignment using DRL}

After the first stage, the camera is close to the peg, losing the peg's direct view, as shown in Figure \ref{fig:partial_obs}. This imposes a challenge when aligning the centers of the gear and peg, as the center of the peg is not directly observable. We train DRL agents to infer the underlying relative positions to the peg's center by observing its neighboring images to address this partial observable problem. 

The process can be formulated as a Partially Observable Markov Decision Process (POMDP) with $\mathcal{M} = \{\mathcal{S}, \mathcal{A}, P, \Omega, \mathcal{T}, R,\gamma\}$. 
In the POMDP,  
$\mathcal{S}$ represents a set of states, 
$\mathcal{A}$ a set of actions, $P: \mathcal{S} \times \mathcal{A} \times \mathcal{S} \mapsto \mathbb{R}$ the state-transition probability function indicating the probability of a state-action pair leading to a specific next state.
$\Omega$ stands for the set of possible observations that the agent can receive,
and $\mathcal{T}: \Omega \times \mathcal{A} \times \Omega \mapsto \mathbb{R}$ is the probability of receiving observation $\Omega$ given that the agent takes action $a$ in state $s$.
The reward function $R : \mathcal{S} \times \mathcal{A} \times \mathcal{S} \mapsto \mathbb{R}$
maps a state-action-state triple to a real-valued reward.
The discount factor $\gamma \in [0, 1]$ controls the relative importance of immediate and future rewards. The goal in DRL is to find a policy $\pi: \mathcal{S} \mapsto \mathcal{A}$, mapping a state to an action that maximizes the expected return from step $t$ 
\begin{equation}
    G_t = \sum_{i=t}^{\infty} {\gamma^{i-t} R(s_i,~a_i,~s_{i+1})}.
\end{equation}
Specifically, in the POMDP, the observation of the agent at time $t$ is the neighbor RGB image of the platform, as shown in Figure \ref{fig:partial_obs}, and action is a two-dimensional offset $\{x_{\delta},~y_{\delta}\}$ to the end-effector's position in $\{x,~y\}$ plane. To find a policy that maximizes the return, we explore DQN \cite{mnih2015human} algorithm for discretized action output and PPO \cite{ppo} for continuous action output. We design a convolutional neural network (CNN), as shown in Figure \ref{fig: network}, to extract features from RGB images and output the movement offset  $\{x_{\delta},~y_{\delta}\}$ actions for DQN or PPO. 

DQN is a value-based approach, and CNN stands for the $Q$ value network. The action is selected by applying the $\arg\max$ function on Q values output on different actions $a_i = \arg\max(Q(s_i, A))$, where $A$ is the movement in $\{x,~y\}$ direction associated with discretized step size $\{-1,~+1,~-5,~+5\}$~\si{\mm}. In contrast, PPO is a policy-based approach, and the CNN represents the policy network that directly maps the image observation into continuous $\{x_{\delta},~y_{\delta}\}$ output in $[-5,~+5]$~\si{\mm}. For more details about DQN and PPO, we refer readers to the literature \cite{mnih2015human} and \cite{ppo}.

\begin{figure}[t]
    \centering
    \subfloat[Scanned platform]{\includegraphics[width=0.45\columnwidth]{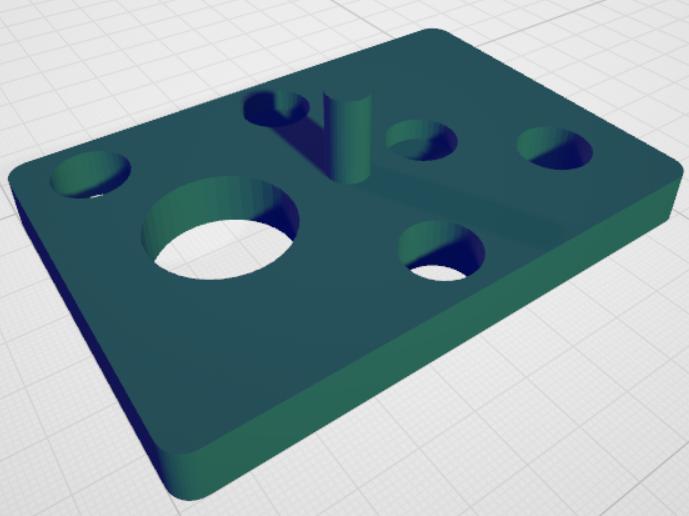}\label{fig:scan}} 
    \hspace{0.05cm}
    \subfloat[Synthetic RGB image]{\includegraphics[width=0.45\columnwidth]{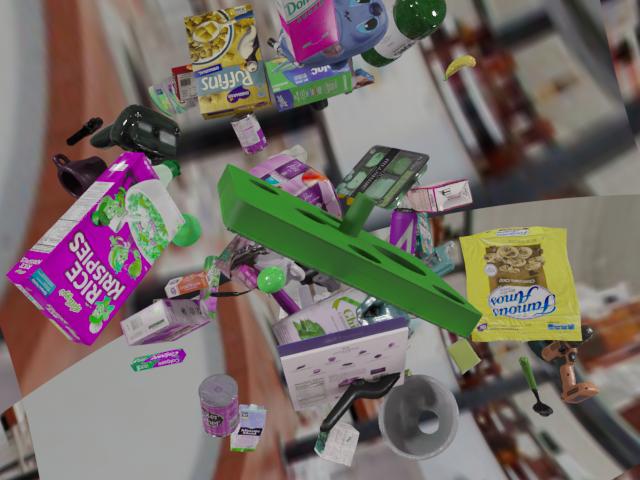}\label{fig:synthetic}}
\caption{The figure displays the 3D mesh of the platform and the synthetic image for training YOLO.}
        \vspace*{-0.27cm}%

\end{figure}

\subsubsection{Training}
The training of deep neural networks is non-trivial due to the high demand for training data. To ease the data collection effort for training YOLO, we use a 3D scanner to obtain the 3D mesh of the target platform, shown in Figure \ref{fig:scan}, and import it into the Blender to generate synthetic images, shown in Figure \ref{fig:synthetic}, with annotations by leveraging the pipeline proposed in \cite{blender_cao}. After training, the pre-trained YOLO is directly deployed in the first stage for coarse localization without further training.

\begin{figure}[t]
    \centering
    \includegraphics[width=6cm]{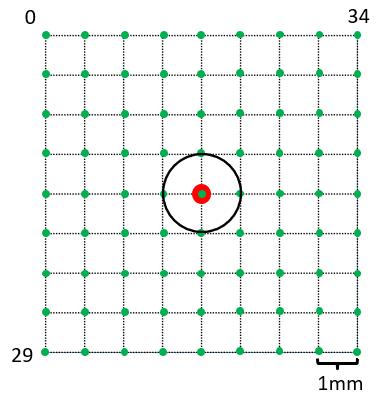}
    \caption{The plot displays a grid world consisting of cells with a length of 1mm, where the center of the peg is marked in red and green grid points indicating the position for capturing real images using the on-wrist camera.}
    \label{fig: grid_data}
    \vspace*{-0.27cm}%
\end{figure}

The training of DRL agents involves extensive interaction with the environment, which imposes a challenge when training in the real world due to the sampling complexity \cite{ibarz2021train}. To address this problem, we discretize the $\{x,~y\}$ plane around the peg as a grid world. We then drive the robot to take pictures at each grid point and record its coordinates, which will be used to construct an environment proximity to interact with the DRL agents for training. As shown in \ref{fig: grid_data}, the grid is a \SI{1}{\mm} x \SI{1}{\mm} square, and the entire map has 30 rows and 35 columns, covering a \SI{3}{\cm} x \SI{3.5}{\cm} space. This sampling process can be automated by programming the movement of the robot without measuring the position of the peg, and it can be done efficiently within 30 minutes.

\begin{figure}[t]
    \vspace*{0.17cm}%

    \centering
    \includegraphics[width=\columnwidth]{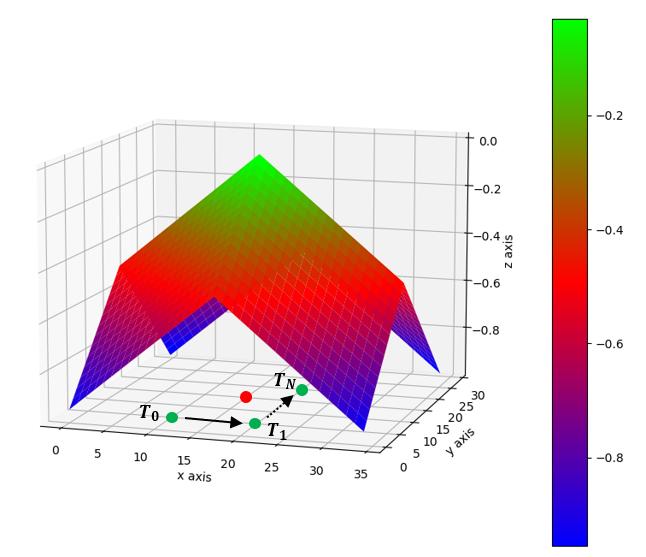}
    \caption{The 3D diagram shows the rewards associated with the training environment, with a red dot marking the center of the peg. The offline training process is visualized using green dots and arrows in the $\{x,~y\}$ plane.}
    \label{fig: pre_train}
            \vspace*{-0.27cm}%

\end{figure}

The grid world enables us to derive a dense reward function to guide the agents to learn more efficiently. As shown in Equation \ref{formula: reward}, The first term of the reward function is the normalized Manhattan distance between the current grid point and the center of the peg, marked as red, where $S_{px}$ and $S_{py}$ are the x,~y coordinate of the grid point, $n_{Y}$ and $n_{X}$ are the total number of rows and columns in the sampling map. The second term is the penalty term to penalize the robot moving out of the \SI{3}{\cm} x \SI{3.5}{\cm} space. The agent receives the highest reward, $r=1$, when the center of the gear reaches the target and receives an additional penalty, $\beta = 10$, when the robot moves out of the boundary, as shown in Figure \ref{fig: pre_train}.
\begin{align}
r=-(\frac{\left | S_{px}  -\frac{n_{X}}{2} \right |}{n_{X}}+\frac{\left | S_{py} -\frac{n_{Y}}{2} \right |}{n_{Y}})/2 - \beta, 
\label{formula: reward}
\end{align}

Now we are ready to introduce the pre-training process. The training episode starts from $T_0$ at a random point $S_{px_0,~py_0}$ in the grid world, as shown in Figure \ref{fig: pre_train}. The observation for the agent is the recorded real-world image from the on-wrist camera. The agents take the image as input and output the movement offset to the current position of the end-effector. By applying the movement offset, the training environment will evolve to the next state at $T_1$ at $S_{px_1,~py_1}$. The interaction continues until the episode reaches the maximum length or the episode terminates when the robot moves out of the grid world. To note, PPO algorithm outputs continuous action which might not lead the next state reaching the exact grid point on the map. In this case, we take the nearest grid point to the center of the end-effector as the next state. This approximation will introduce a gap between the pre-training environment and real world, which will be mitigated through fine-tuning in the real world.

After pre-training, we transfer the learned agents to the real world and further train DRL agents with sparse reward $[0, 1]$ to compensate for the inaccuracy of the pre-training environment to reach optimal. 

\section{Experiments and results}
\label{sec:experiment}
\begin{figure}[h]
    \vspace*{0.17cm}%
    \centering
    \includegraphics[width=0.9\columnwidth]{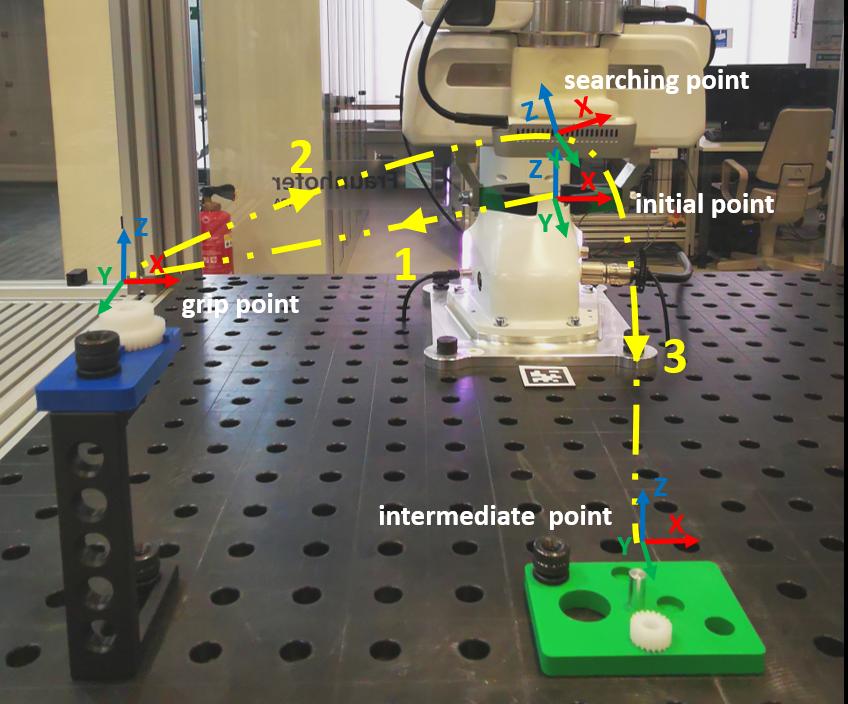}
    \caption{The figure depicts the workflow of gear assembly, consisting of three steps: 1) driving the robot to grasp the gear on the placement tray; 2) returning the robot to the initial position while holding the gear; and 3) completing the assembly process using the proposed two-stage approaches.}
    \label{fig: trajectory}
\end{figure}

\begin{figure}[htbp]
    \centering
    \includegraphics[width=0.95\columnwidth]{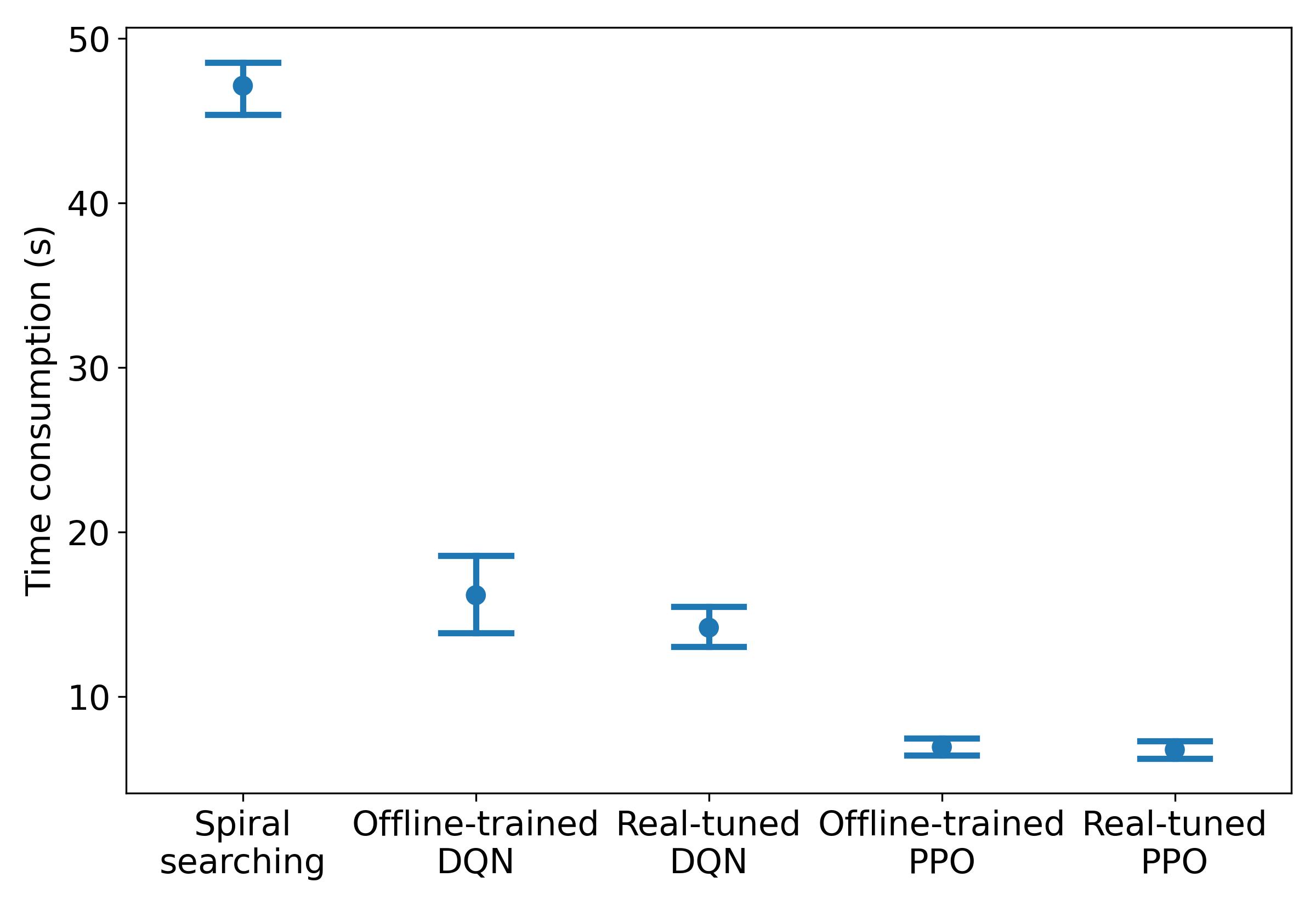}
    \caption{The plot illustrates the run time performance of various approaches in gear assembly experiments with 100 different initial starting points, where the confidence interval is \SI{95}{\percent}. }
    \label{fig: box}
\end{figure}

\begin{figure*}[ht]
  \vspace*{0.17cm}%
  \centering
  \hspace{0.05cm}
  \subfloat[Spiral Searching]{\includegraphics[width=0.29\textwidth]{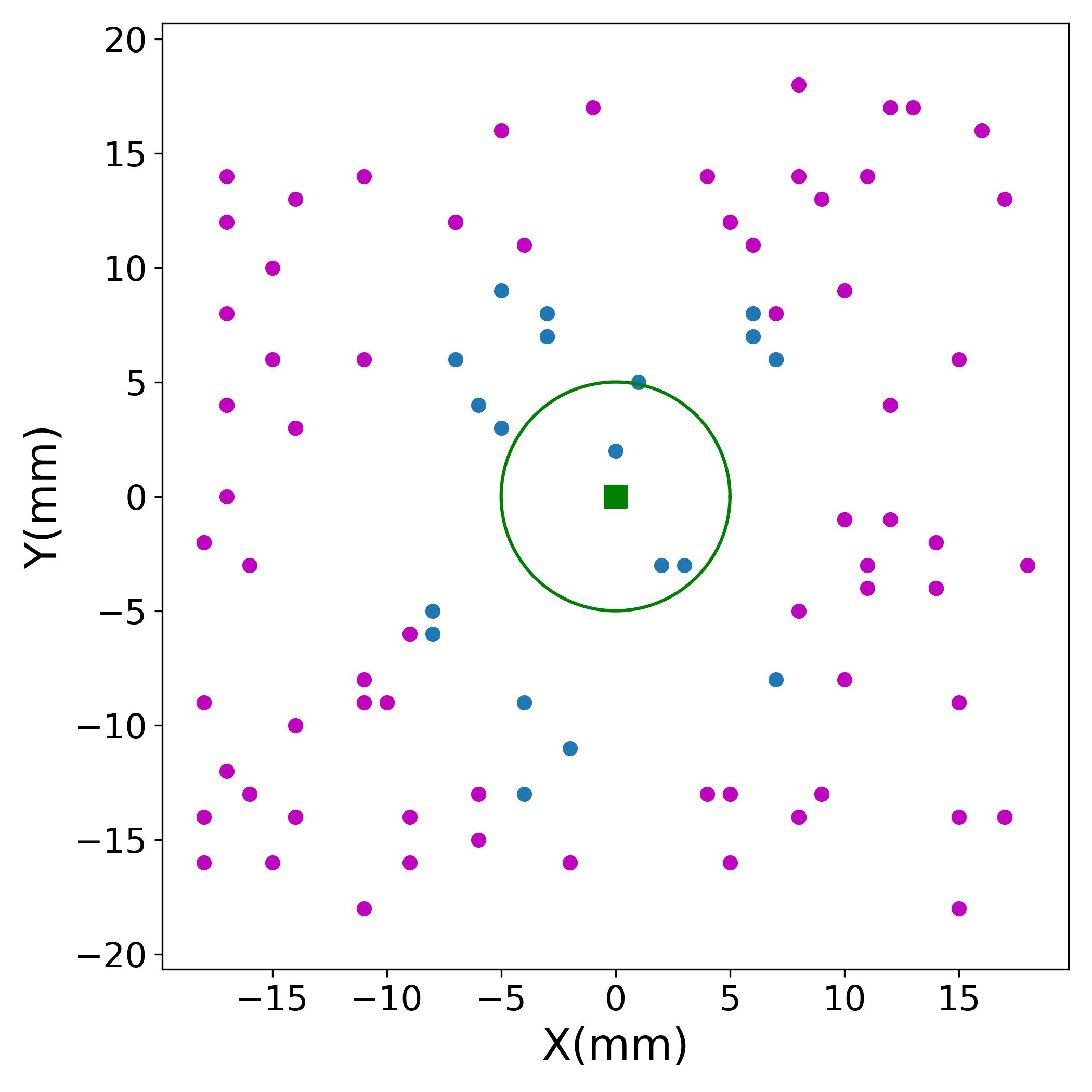}\label{fig:rt_1}} 
  \subfloat[Offline-trained DQN]{\includegraphics[width=0.29\textwidth]{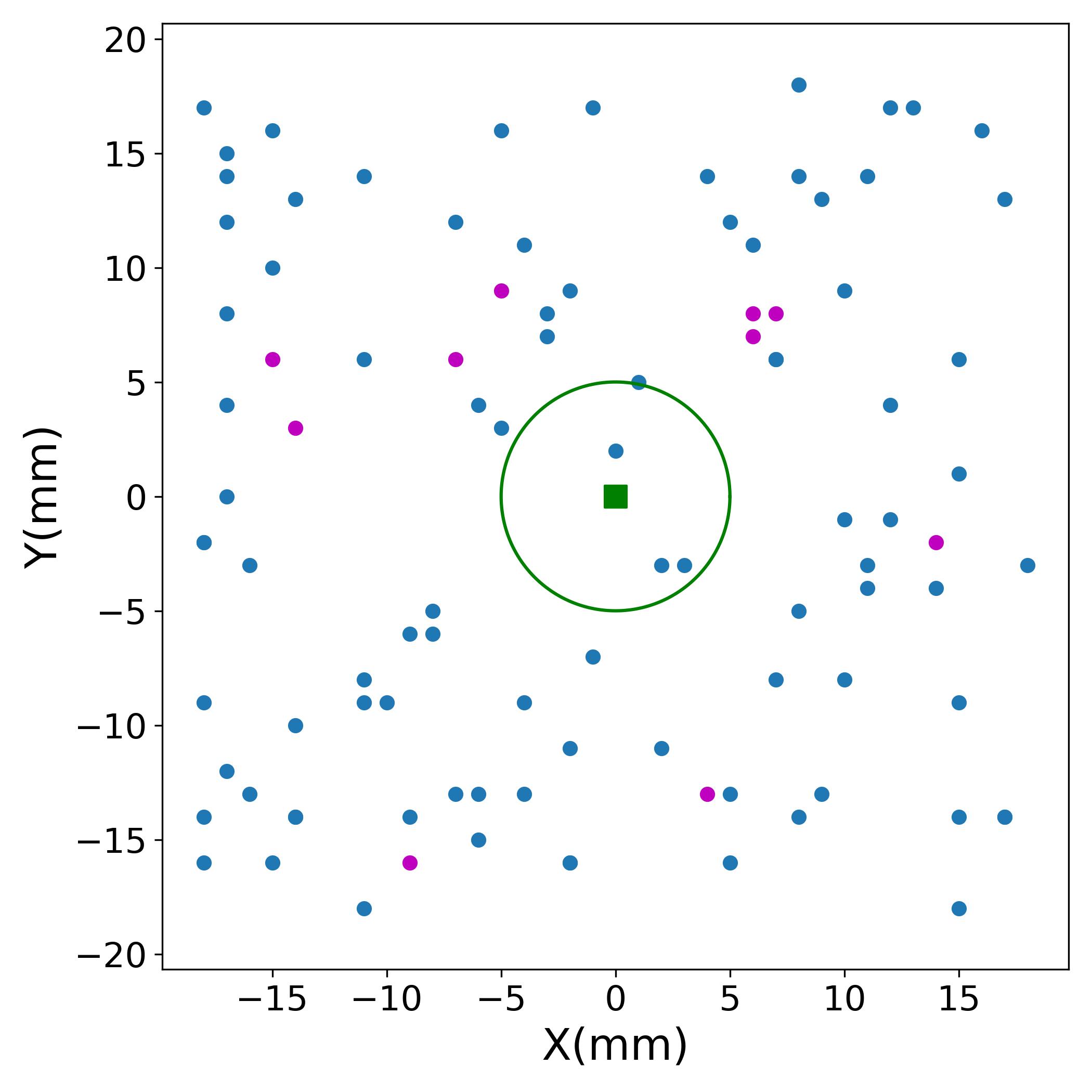}\label{fig:rt_3}}
  \subfloat[Offline-trained PPO]{\includegraphics[width=0.29\textwidth]{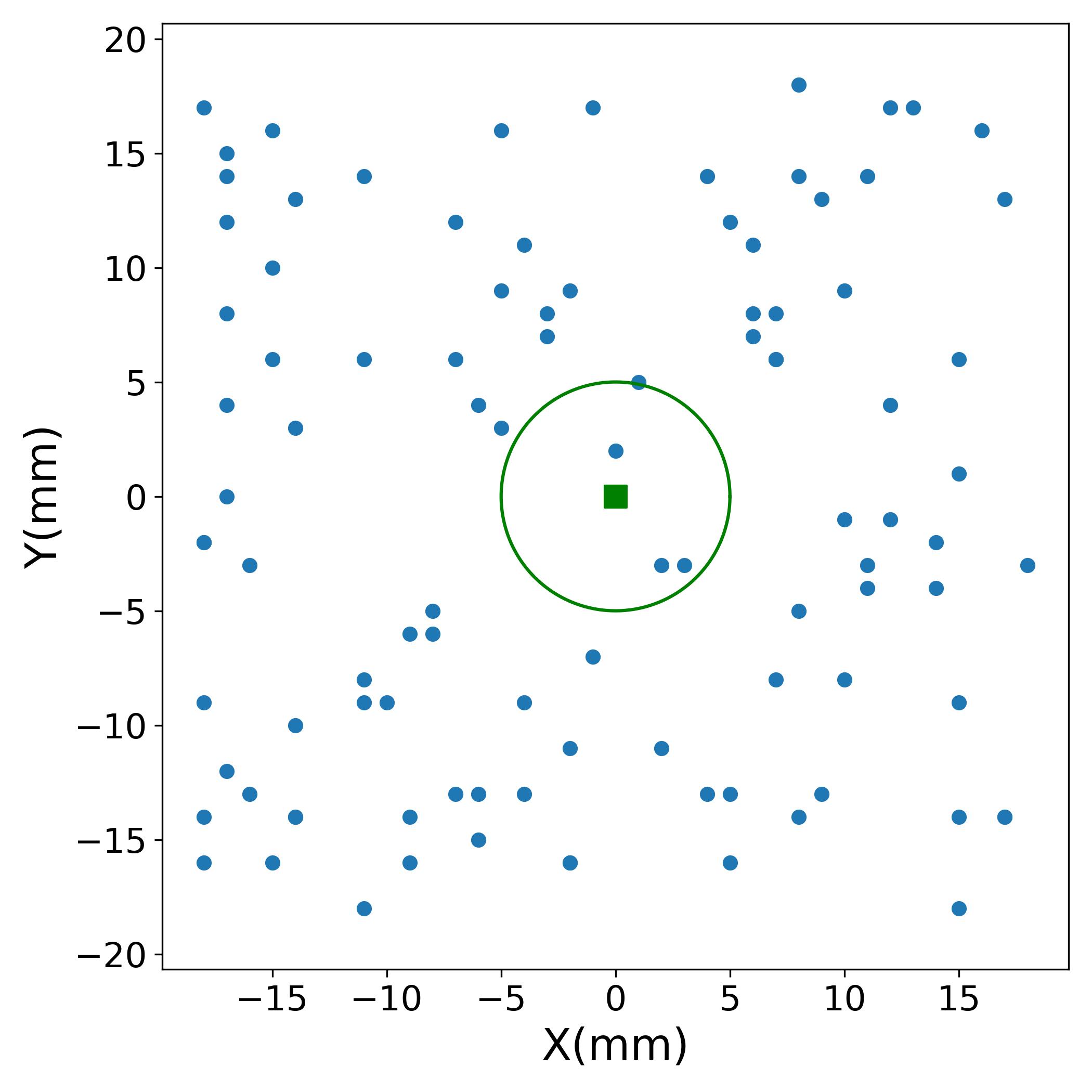}\label{fig:rt_4}}
  \subfloat{\includegraphics[width=0.12\textwidth]{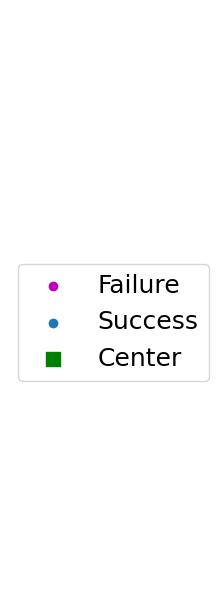}} 
  \caption{The plot demonstrates the robustness performance of different approaches in gear assembly experiments with 100 various initial starting points, where blue dots represent successful assemblies, purple dots denote failure cases and the green circle corresponds to the shape of the peg.}
  \label{fig: robustness_test}
\end{figure*}

In this section, we introduce the experimental setup of the assembly task and the evaluation of the proposed methods.
\subsection{Hardware setup}

The assembly scenario consists of three main components: a placement tray, a Frank Emika Panda Robot, and the target platform, as shown in Figure \ref{fig: trajectory}. The placement tray is a C-shaped bracket with a blue platform on top for initial gear placement. The target platform has an aluminum peg at the center and a gear mounted for gear assembly with a tolerance of \SI{0.3}{\mm} \footnote{The two gears used in this work are GEABP1.0-40-10-B-10 and GEABP1.0-20-10-B-10 respectively.}. The robot is embedded with a Realsense D415 RGBD camera on the wrist to observe the environment and a force sensor on the gripper to detect contact. Moreover, the robot's gripper fingers are customized to be a pair of crabs to increase grasping reliability.

As the initialization, the gear is placed at the placement tray horizontally at a pre-configured known position to reduce the grasping uncertainty. The target platform is mounted horizontally on the workbench. The robot arm is initialized to view the target platform at any position of the $\{x,y\}$ plane and above a certain height. As depicted in Figure \ref{fig: trajectory}, The workflow of the whole pipeline includes 1) the robot is programmed to grasp the gear on the placement tray; 2) the robot returns to the initial position with the gear in hand; 3) the assembly starts using our proposed two-stage approach. 

\subsection{Software setup}

In the proposed two-stage assembly approach, we use the Darknet implementation of YOLO as in \cite{yolov3} for object detection and further integrate it with ROS framework as in \cite{yolo_ros} for efficient data communication. We leverage the off-the-shell implementation for PPO and DQN from \cite{stable-baselines3} to address the center alignment problem at the second stage. The robot's motion control and contact detection are performed by \textit{pitasc} \cite{pitasc}, a modular robot operating system. Furthermore, We use ROS as a communication bridge between different modules.

\subsection{Evaluation}
The target localization using YOLO in the first stage can efficiently and reliably bring the in-hand gear to the neighbor of the peg, leaving the center alignment in the second stage as an essential factor for successful assembly. Therefore, we mainly evaluate the performance of the second stage, in which we initialize the second stage at randomly sampled 100 points in the neighbor of the peg and study the robustness and running efficiency of PPO and DQN. Furthermore, we implement a heuristic spiral search approach as a comparison baseline. 

\subsubsection{Robustness performance}
In the robustness test, we are interested in answering how reliable different approaches can achieve in assembly tasks with different starting points. We set the maximum run time budget as 50 seconds, and the experiment that can finish assembly within the budget will be treated as a success. We repeat the experiment 100 times from different randomly sampled starting points and report a success rate (SR) for spiral search and offline-trained DQN and PPO approaches. We furthermore report the SR of DQN and PPO after real-world fine-tuning.

As shown in Figure \ref{fig: robustness_test}, the offline-trained PPO achieves the best performance with 100\% SR, whereas the offline-trained DQN fails in many cases, as marked as purple, resulting in a 90\% SR. The reason is that the relative pose, i.e., a relative rotation or translation offset, of the target platform to the end-effector is not perfectly calibrated before sampling, which introduces a mismatch regarding the origin and axis in the constructed offline training environment. This mismatch imposes a big challenge for DQN, as DQN outputs discretized action along the ${x, y}$ axis. The policy learned in the offline environment might not have a valid solution to align the centers in the real world finely. However, PPO outputs continuous actions, which has more compliance that could compensate for planning errors in the real world. Spiral searching only has a 19\% SR, showing its sensitivity to the distance to the peg, which also implies the demand for the precision of the target localization.

We further fine-tune the learned DQN and PPO in the real world to study the gap between the offline training environment and the real world. The real-world fine-tuning takes approximately 10 minutes, after which DQN can achieve a 100\% SR.

\begin{figure*}[h!]
  \centering
  \subfloat[Spiral searching]{\includegraphics[width=0.32\textwidth]{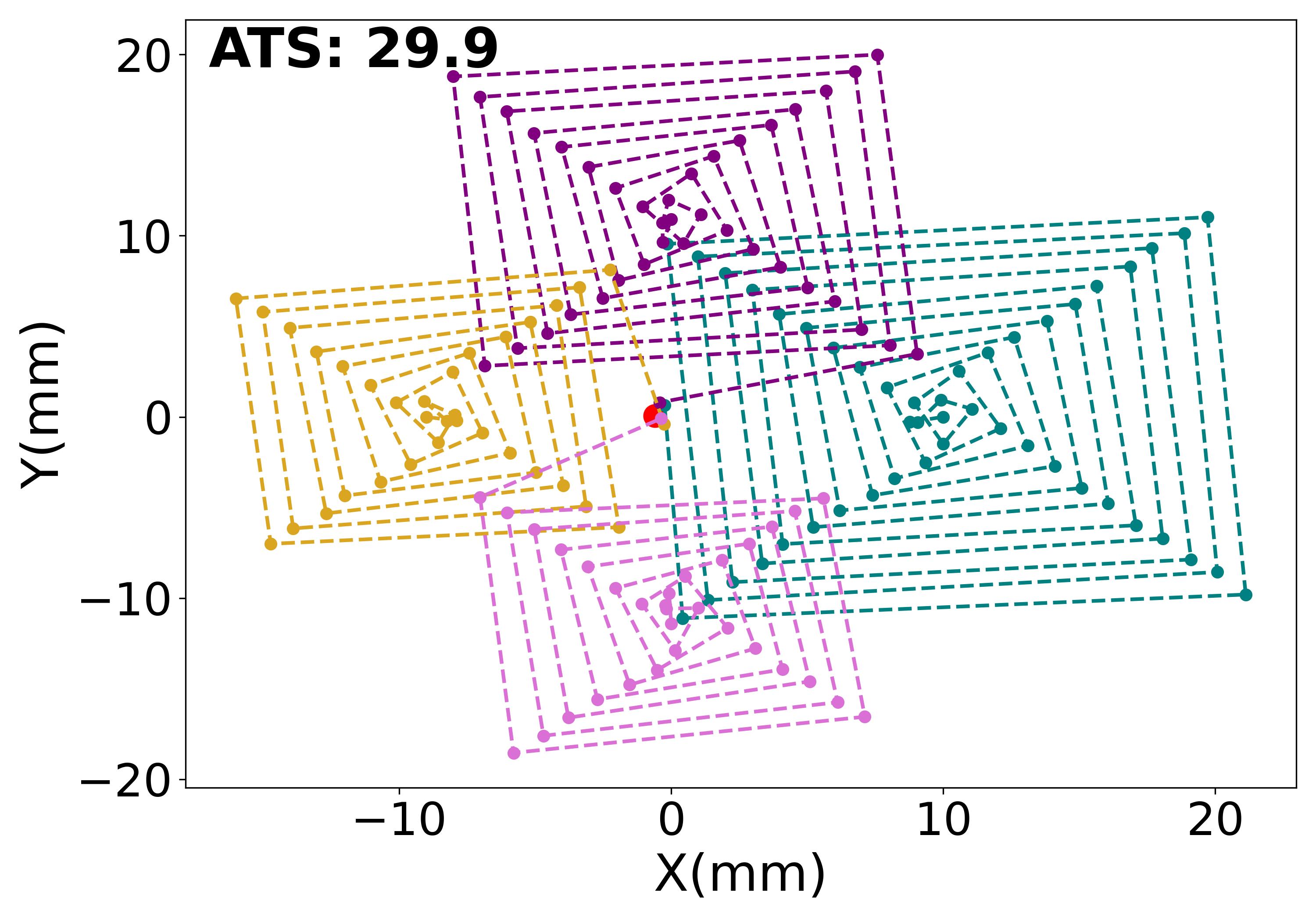}\label{fig:pt_1}}
  \hspace{0.05cm}
  \subfloat[Real-world tuned DQN]{\includegraphics[width=0.32\textwidth]{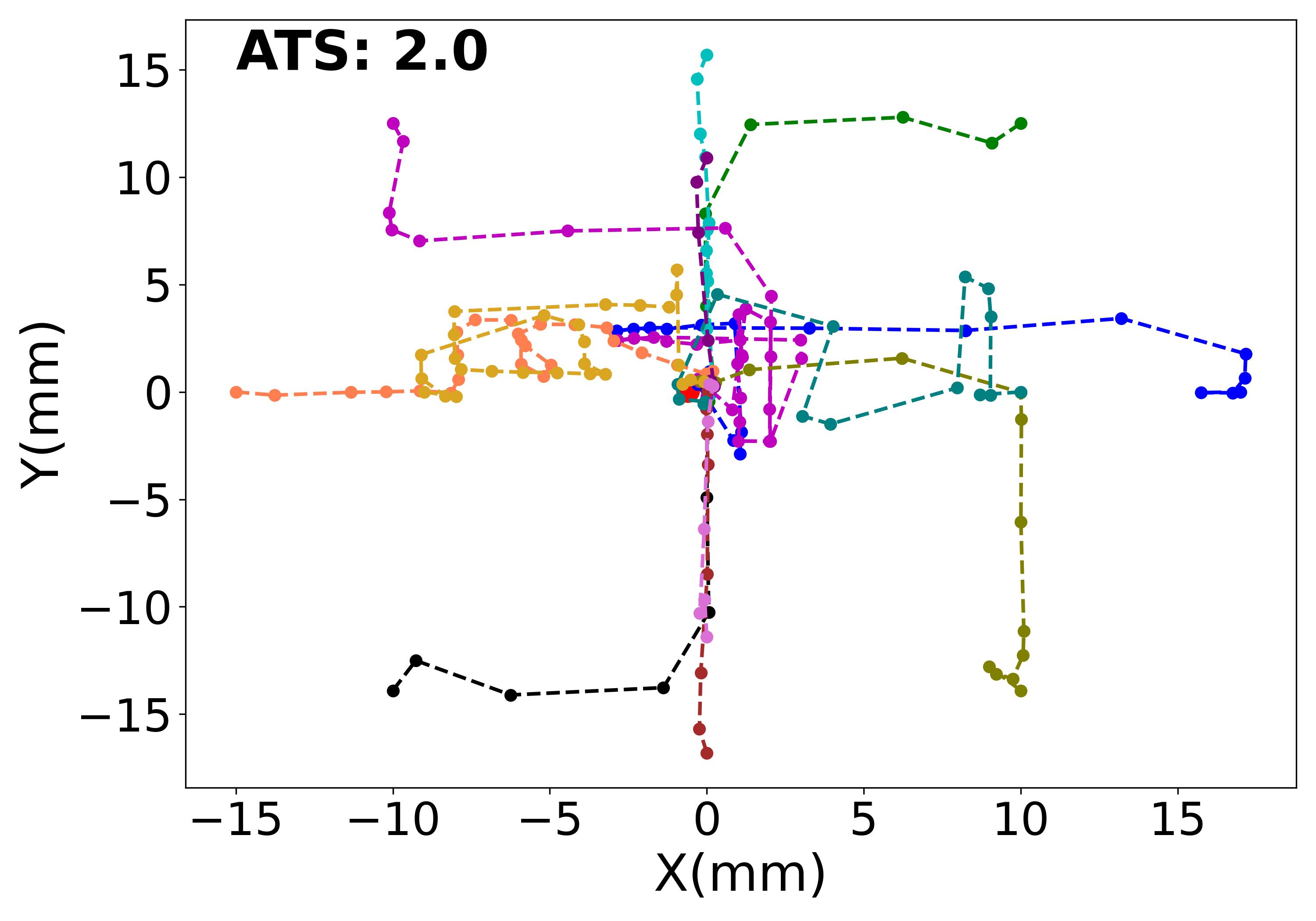}\label{fig:pt_2}}
  \hspace{0.05cm}
  \subfloat[Real-world tuned PPO]{\includegraphics[width=0.32\textwidth]{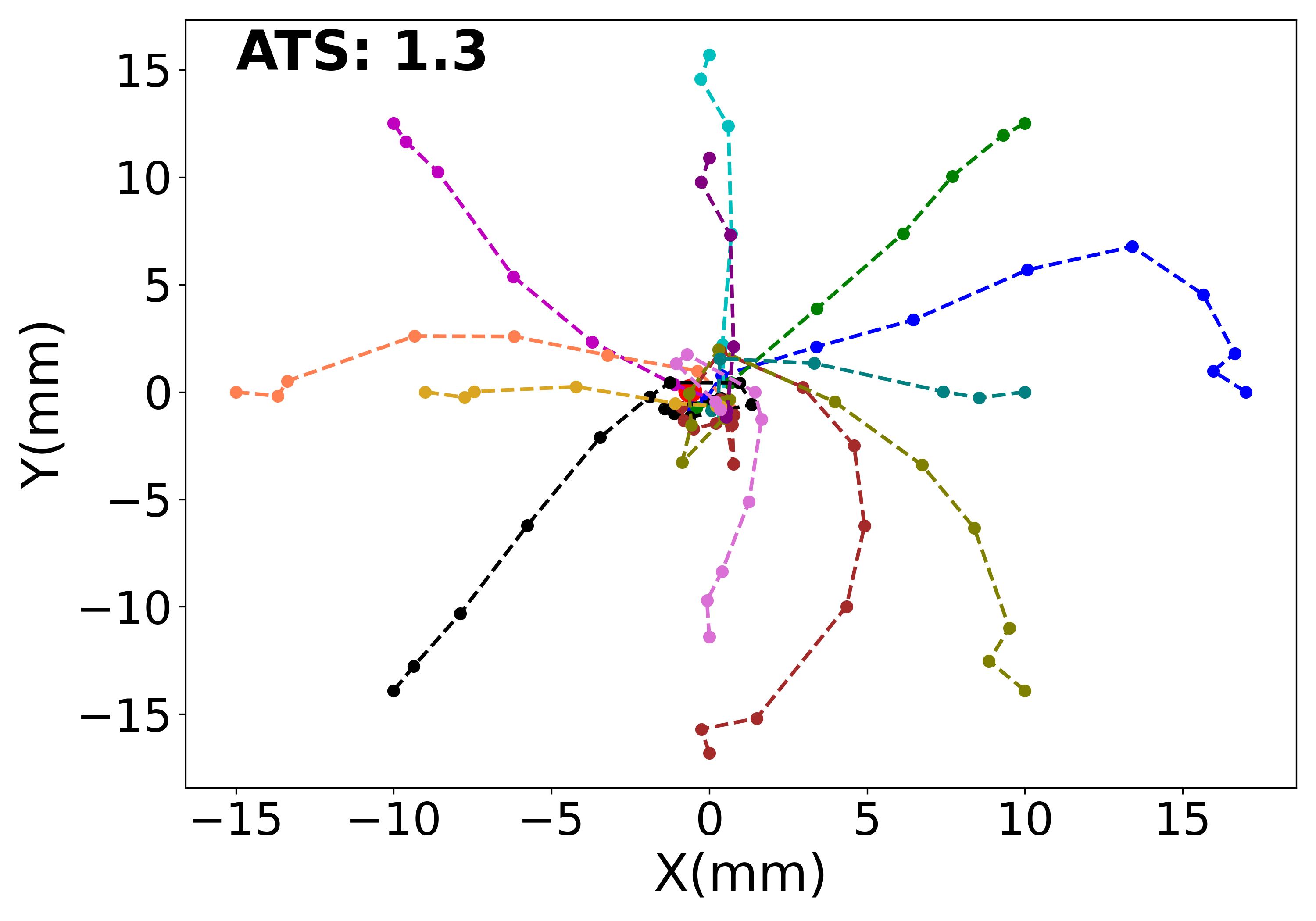}\label{fig:pt_3}}
  \caption{The figure displays the trajectories of gear assembly from various initial starting points using different methods, where PPO achieves the most efficient performance with an average traveling score (ATS) of 1.3.}
  \label{fig: efficiency_test}
\end{figure*}


\subsubsection{Efficiency performance}
In the efficiency test, we aim to study how fast different approaches can complete the assembly task. We additionally record the time consumed for each experiment performed in the robustness test, where failed experiments consume a maximum time of 50 seconds.

As shown in Figure \ref{fig: box}, the real-world tuned PPO is the most efficient algorithm that takes approximately 6 seconds to finish the second stage, whereas DQN, on average, takes 14 seconds. Spiral searching fails in cases where starting points are far away from the center of the peg, resulting in the most inefficient solution. Real-world tuning is generally beneficial to improve performance consistency and reduce the average run time, especially for DQN. However, we also notice that the policies learned from the offline training environment have comparable performance to the real-world tuned policies. This indicates that learning only in the offline environment can achieve sufficiently good efficiency, showing the potential of training policies solely in the offline-environment constructed with sampling points under tolerable accuracy.

To analyze the behaviors of different approaches, we test spiral search, real-world tuned DQN and PPO from 12 different starting points across the $\{x,y\}$ plane, and their trajectories are visualized in Figure \ref{fig: efficiency_test}. The Figure shows that the trajectories of the PPO are closer to the optimal trajectories, i.e., straight line between two points. In contrast, the DQN performs in Manhattan geometry due to the discretized action space. To quantitatively study the difference, we compute the traveling score $T_{score}$ for each trajectory as 

\begin{equation}
\hfill T_{score} = \frac{L_{traj}}{|P_{target} - P_{start}|}\hfill,
\label{formula: traveling_scale}
\end{equation}

where $L_{traj}$ is the actual length of the trajectory,  $|P_{target} - P_{start}|$ stands for the euclidean distance between the start and end point. The optimal trajectory would lead to a score of 1. As shown in Figure \ref{fig: efficiency_test}, PPO achieves the average traveling score (ATS) of 1.3 among 12 trajectories, and DQN achieves 2.0. Spiral searching only succeeds in the four nearest starting points, and the ATS for these four trajectories is 29.9. Although PPO acts close to optimal solutions, it can perform a curved trajectory starting from the bottom right corner, as shown in Figure \ref{fig:pt_3}. This is possibly due to the space near the bottom right corner is not well explored during training. This can be mitigated by resetting the training from this region or using more exploration-efficient off-policy algorithms. 

With the most efficient real-world tuned PPO, the whole assembly pipeline can be completed within approximately 15 seconds, in which the first stage takes roughly 9 seconds to finish object searching and approaching.
\section{CONCLUSIONS}
\label{sec:conclusion}
In conclusion, we presented a novel two-stage approach for autonomous assembly that integrates YOLO for target coarse localization and deep reinforcement learning for fine alignment. Our proposed approach exhibits impressive robustness and running efficiency when evaluated in an industrial gear assembly task with 0.3mm precision tolerance. This has been possible thanks to the use of deep reinforcement learning that can effectively address the problem of partial visibility when the on-wrist camera is too close to the target and the force feedback to simplify the transitions between two stages. We also introduced a novel offline training strategy using real-world data to pre-train DRL agents.

The limitation of this work is that it requires visually distinguishable spatial features on the part that can signify its relative distance from the background. In other words, our approach would struggle to learn from a broad plane of uniform pattern and color, where every observation at different positions looks identical. To overcome this limitation, one can, for example, place printed irregular pattern near the part to enhance spatial information. The current work is tested only in the 3D space with simplified grasping. Future work will extend to complete the whole pipeline by combining the grasping and assembly in 6D space.


\section*{ACKNOWLEDGMENT}
Marco Caccamo was supported by an Alexander von Humboldt Professorship endowed by the German Federal Ministry of Education and Research.

The authors would like to thank Lukas Dirnberger for his assistance in synthetic data generation.
 
\balance
\bibliographystyle{ieeetr}
\bibliography{ref}
\end{document}